\documentclass{article}

\usepackage{arxiv}

\usepackage[utf8]{inputenc} 
\usepackage[T1]{fontenc}    
\usepackage[hidelinks]{hyperref}       
\usepackage{url}            

\usepackage{authblk}
\usepackage{booktabs}       

\usepackage{amsfonts}       
\usepackage{nicefrac}       
\usepackage{graphicx}
\usepackage{doi}
\usepackage{float}
\usepackage{stfloats}
\usepackage{subcaption}
\usepackage{textcomp, gensymb}
\usepackage{enumitem}
\usepackage{multirow}
\usepackage{xcolor}
\usepackage{pgfplots}
\usepackage{wrapfig}
\usepackage{rotating}
\usepackage{todonotes}
\usepackage[ruled]{algorithm2e}
\usepackage{array}
\newcolumntype{x}[1]{>{\centering\arraybackslash\hspace{0pt}}p{#1}}

\pgfplotsset{compat=1.18}

\usepackage{tikz}
\usepackage{tikz-qtree}
\usetikzlibrary{trees}
\usetikzlibrary{mindmap}
\usetikzlibrary{positioning}

\begin{document}


\title{A Domain-Based Taxonomy of Jailbreak Vulnerabilities in Large Language Models}

\newcommand{\shorttitle}{A taxonomy of jailbreaking of Large Language Models through domain characterization} 

\author{Carlos Pel\'aez-Gonz\'alez$^{*}$, Andr\'es Herrera-Poyatos, Cristina Zuheros, David Herrera-Poyatos, Virilo Tejedor, Francisco Herrera}

\affil{Department of Computer Science and Artificial Intelligence, Andalusian Institute of Data Science and Computational Intelligence (DaSCI), University of Granada, Spain. \\ $^{*}$Corresponding author. Emails: \texttt{\{carlosprog, andreshp, czuheros, divadhp\}@ugr.es}, \texttt{virilo@gmail.com}, \texttt{herrera@decsai.ugr.es}}

\date{\today}

\maketitle              

\begin{abstract}

The study of large language models (LLMs) is a key area in open-world machine learning. Although LLMs demonstrate remarkable natural language processing capabilities, they also face several challenges, including consistency issues, hallucinations, and jailbreak vulnerabilities.  Jailbreaking refers to the crafting of prompts that bypass alignment safeguards, leading to unsafe outputs that compromise the integrity of LLMs. This work specifically focuses on the challenge of jailbreak vulnerabilities and introduces a novel taxonomy of jailbreak attacks grounded in the training domains of LLMs. It characterizes alignment failures through generalization, objectives, and robustness gaps. 

Our primary contribution is a perspective on jailbreak, framed through the different linguistic domains that emerge during LLM training and alignment. This viewpoint highlights the limitations of existing approaches and enables us to classify jailbreak attacks on the basis of the underlying model deficiencies they exploit.

Unlike conventional classifications that categorize attacks based on prompt construction methods (e.g., prompt templating), our approach provides a deeper understanding of LLM behavior. We introduce a taxonomy with four categories ---mismatched generalization, competing objectives, adversarial robustness, and mixed attacks--- offering insights into the fundamental nature of jailbreak vulnerabilities. Finally, we present key lessons derived from this taxonomic study.

\end{abstract}
\section{Introduction} 

Large Language Models (LLMs) have significantly transformed the AI landscape in recent years. Originally designed to predict word sequences based on given inputs~\cite{zhao_survey_2023}, LLMs leverage the transformer architecture and vast amounts of training data. Due to their emergent capabilities, these models can perform various natural language processing tasks without the need for retraining or fine-tuning~\cite{wei_emergent_2022}. To ensure that LLM outputs align with human values and ethical standards, model alignment has been proposed as a crucial step in their development~\cite{ouyang_training_2022}.

Despite their capabilities, LLMs face several challenges, including consistency issues, hallucinations, and model jailbreaks~\cite{yao_survey_2024}. In this work, we focus on the latter. Model jailbreak refers to the act of bypassing safety mechanisms through techniques including prompt engineering, leading the model to generate unsafe or unintended outputs despite the presence of security guardrails. Such vulnerabilities can compromise user safety, erode trust in AI systems, violate regulatory standards, and propagate misinformation~\cite{chen_combating_2023}. Therefore, mitigating the impact and success rate of jailbreak attacks is essential when developing LLMs.

Existing defenses against model jailbreaks primarily focus on detecting unsafe queries or responses, refining model alignment algorithms, and enhancing the quality of alignment datasets through adversarial testing (red-teaming)~\cite{yao_survey_2024}. These methods aim to ensure that LLMs adhere to ethical and safety guidelines, even when faced with adversarial inputs. However, despite these efforts, novel jailbreak techniques continue to emerge, effectively circumventing existing safeguards~\cite{deng_masterkey_2024,handa_jailbreaking_2024,anil_many-shot_2024,hayase_query-based_2024}. This ongoing evolution of jailbreak strategies poses a persistent challenge, as attackers continuously discover new ways to exploit model vulnerabilities and undermine the effectiveness of current defenses.

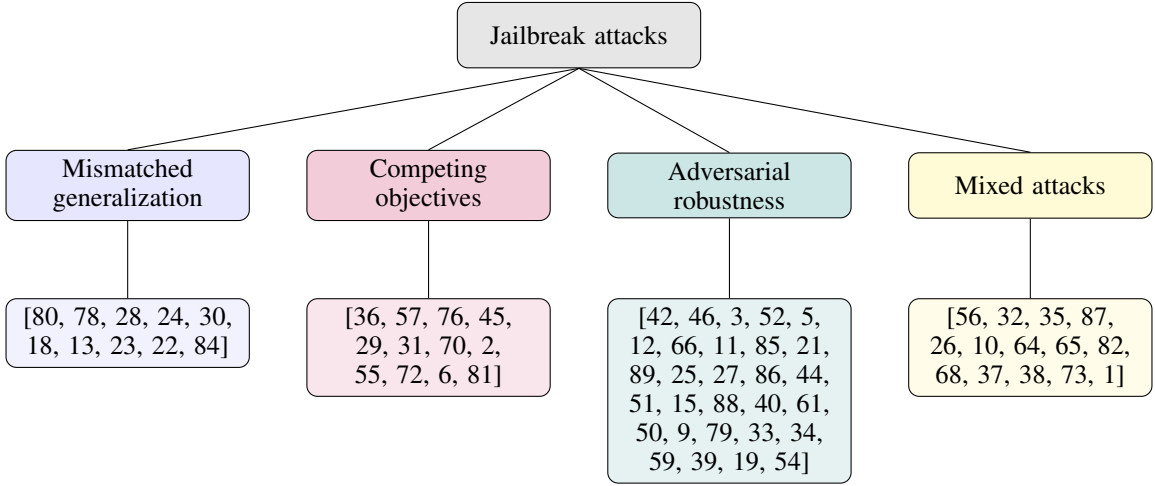
\begin{figure}
    \centering
    \begin{tikzpicture}[
            level 1/.style={sibling distance=40mm, level distance=20mm, text width=30mm},
            level 2/.style={sibling distance=40mm, level distance=15mm, text width=30mm, anchor=north}, 
            frontier/.style={distance from root=48mm},
            every node/.style={rectangle, rounded corners, draw=black, align=center},
        ]
        \node [text width=30mm, fill=gray!20, minimum height=8.7mm] {Jailbreak attacks}
        child {
            node [fill=blue!10] (mg) {Mismatched generalization}
            child { node [fill=blue!5] {\cite{yuan_gpt-4_2024, yong_low-resource_2023, lemkin_using_2024, jiang_artprompt_2024, li_cross-language_2024, handa_when_2024, ghanim_jailbreaking_2024, jeong_playing_2024, huang_catastrophic_2024, zhao_weak--strong_2024}} }
        }
        child {
            node [fill=purple!20] {Competing objectives}
            child { node [fill=purple!10] {\cite{liu_jailbreaking_2023, shen_anything_2024, yao_fuzzllm_2024, perez_ignore_2022, li_multi-step_2023, li_deepinception_2024, wei_jailbreak_2023, anil_many-shot_2024, shah_scalable_2023, wu_jailbreaking_2024, chao_jailbreaking_2023, zeng_how_2024}} }
        }
        child {
            node [fill=teal!20] {Adversarial robustness}
            child { node [fill=teal!10] {\cite{niu_jailbreaking_2024, qi_visual_2024, bagdasaryan_abusing_2023, schlarmann_adversarial_2023, bailey_image_2024, gao_adversarial_2024, wang_white-box_2024, dong_how_2023, zhao_evaluating_2023, hu_transferable_2024, zou_universal_2023, jones_automatically_2023, lapid_open_2023, zhu_autodan_2024, paulus_advprompter_2024, sankar_sadasivan_fast_2024, guo_cold-attack_2024, zhuo_robustness_2023, mehrotra_tree_2024, takemoto_all_2024, samvelyan_rainbow_2024, deng_masterkey_2024, yu_llm-fuzzer_2024, liu_autodan-turbo_2024, liu_autodan_2024, sitawarin_pal_2024, maus_black_2023, hayase_query-based_2024, shah_loft_2023}} }
        }
        child {
            node [fill=yellow!20, minimum height=8.7mm] {Mixed attacks}
            child { node [fill=yellow!10] {\cite{shayegani_jailbreak_2024, li_images_2024, liu_arondight_2024, zhuo_red_2023, kang_exploiting_2023, ding_wolf_2024, wang_robustness_2023, wang_adversarial_2023, zhang_safety_2023, wei_jailbroken_2023, liu_flipattack_2024, lv_codechameleon_2024, xu_cognitive_2024, andriushchenko_jailbreaking_2024}} }
        };
    \end{tikzpicture}
    \caption{A summarized version of our proposed taxonomy. The complete taxonomy is described on \autoref{sec:taxonomy}}
    \label{fig:taxonomy_sum}
\end{figure}

In this paper, we investigate the challenges of model alignment and analyze the underlying factors that enable jailbreak attacks despite extensive safety measures. We examine how the inherent complexity of aligning models with ethical principles and intended behaviors contributes to persistent vulnerabilities. Additionally, we explore the specific mechanisms through which these weaknesses manifest, identifying critical gaps in current alignment strategies. Our main contributions are as follows:

\begin{itemize}
    \item We provide a concise overview of contemporary research on model alignment, emphasizing key aspects relevant to understanding jailbreak attacks.
    \item Building on~\cite{wei_jailbroken_2023}, we characterize the language domains that emerge during LLM training. This allows us to formally define the primary weaknesses that facilitate jailbreaking: mismatched generalization, competing objectives, and lack of robustness.
    \item We utilize these definitions to systematically classify jailbreak attacks (see Figure~\ref{fig:jailbreak_taxonomy}), identifying which specific vulnerabilities are exploited in different attack methodologies. This structured approach enhances our understanding of jailbreak techniques and aids in developing more effective countermeasures for future LLMs.
\end{itemize}

The remainder of this paper is organized as follows. Section~\ref{sec:alignment_overview} introduces model alignment and briefly discusses existing techniques. Section~\ref{sec:domain_analysis} explores the language domains involved in LLM training and alignment, formalizing the concepts of mismatched generalization, competing objectives, and lack of robustness. In Section~\ref{sec:taxonomy}, we apply these concepts to classify jailbreak attacks. Section~\ref{sec:Lessons} discusses insights derived from our taxonomy, including open challenges. Finally, we conclude our analysis in Section~\ref{sec:conclusions}.

\section{A brief discussion of LLMs alignment to understand jailbreak attacks context} 
\label{sec:alignment_overview}

Model alignment refers to the process of ensuring that a model's behavior aligns with human preferences by adhering to predefined ethical guidelines, values, and intended objectives~\cite{hendrycks_unsolved_2022}. Large language models (LLMs) are typically trained in two main stages: the first, known as generative pre-training, focuses on learning language patterns~\cite{radford2018improving}, while the second stage is dedicated to aligning the model with human expectations and ethical considerations.

During the generative pre-training phase, models are trained on a vast corpus of text using an autoregressive approach. In this method, a sequence of text is truncated at a certain point, and the model is tasked with predicting the next token in the sequence. While this process is technically a form of supervised learning, the input data consists of unstructured text rather than explicitly labeled samples. As a result, generative pre-training is often considered a form of unsupervised learning. To clarify this distinction, the literature classifies this approach as self-supervised training~\cite{touvron_llama_2023}.

Following the generative pre-training phase, the model acquires the ability to predict the next token in any given sequence. However, since a substantial portion of the training data is typically sourced from the Internet, the model may inadvertently learn biases and exhibit toxic behavior~\cite{bai_training_2022}. To mitigate these issues, a fine-tuning phase is introduced to align the model with human preferences. Unlike self-supervised learning, this phase employs preference learning~\cite{ouyang_training_2022}. Rather than minimizing the error between the model’s output and a predefined ground truth, the model is trained to generate responses that are preferred by users. Notably, multiple outputs can be equally preferred, providing the model with greater flexibility during training. Human preferences can be represented in various ways, such as assigning scores to individual samples or ranking pairs of samples based on preference order.

Preference learning is extensively utilized in reinforcement learning~\cite{wirth_survey_2017} and involves leveraging a preference dataset to optimize the policy of AI agents based on a reward function. By incorporating human preferences, learning algorithms guide model behavior to align with these preferences. One of the pioneering works in applying human preferences to complex learning tasks is~\cite{christiano_deep_2017}, where recommendation systems were trained using a dataset in which humans compared and ranked pairs of short videos according to personal preference. A key advantage of preference learning over traditional approaches is its efficiency: it requires significantly smaller datasets and can learn robust reward functions within a timeframe ranging from a few minutes to several hours.


One of the pioneering works in applying reinforcement learning from human preferences to generative pre-trained language models is~\cite{stiennon_learning_2020}. This study employed a reinforcement learning approach similar to~\cite{christiano_deep_2017}, utilizing the Proximal Policy Optimization (PPO) algorithm~\cite{schulman_proximal_2017} to enhance summary generation. OpenAI later extended this idea to GPT-3~\cite{ouyang_training_2022}, incorporating an additional step between generative pre-training and preference learning. This intermediate step consists of a supervised training phase on a dataset of crowdworkers' responses to user prompts, designed to mimic the desired chatbot behavior.

Then, a third training stage is applied, commonly known as alignment stage. This stage uses a curated preference dataset which is structured around three key principles: helpfulness, harmlessness, and honesty. Helpfulness ensures that the model follows human instructions as effectively as possible. Harmlessness dictates that the model should refuse instructions that could result in harm to users or others. Honesty ensures that the model avoids generating factually incorrect information.

An alternative successful approach was developed by the Anthropic team~\cite{bai_training_2022}, where the key distinction lies in the explicit separation of helpful and harmful queries within the dataset. During AI assistant interactions, crowdworkers are assigned different tasks: some select the most helpful and honest response from the AI assistant, while others engage in red teaming—identifying and ranking the most harmful responses. This structured approach facilitates the creation of a well-balanced preference dataset for training more aligned AI models.


For a comprehensive survey on model alignment in large language models (LLMs), we refer to~\cite{wang_aligning_2023}. Here, we highlight a novel optimization approach for model alignment known as Direct Preference Optimization (DPO)~\cite{rafailov_direct_2023}. DPO reformulates reinforcement learning into a mathematically equivalent supervised learning problem by introducing a specific loss function. This method offers two primary advantages. First, it eliminates the need for a separate reward model, preventing potential exploitation of the reward function by the reinforcement learning algorithm. Second, it significantly reduces training time, as reinforcement learning is typically more computationally intensive. Despite its promise, it remains uncertain which alignment approach ---DPO or reinforcement learning--- yields superior safety outcomes~\cite{xu_is_2024,ramamurthy_is_2023}. Both methods present inherent challenges that must be addressed to ensure robust model alignment.

Given the multi-stage training process of large language models (LLMs), which includes pre-training followed by alignment, the next section will examine the challenges associated with alignment, specifically through the perspective of jailbreak attacks.

\section{Characterizing the domains in LLM training: towards understanding the weaknesses of LLMs with respect to jailbreaking} 
\label{sec:domain_analysis}

Despite extensive efforts by the research community to align large language models (LLMs) with human preferences, these models remain susceptible to jailbreak attacks. An LLM is considered to be under attack when an adversary successfully induces harmful behavior by manipulating model parameters, often by crafting a carefully designed input prompt. A successful attack circumvents the alignment safeguards that are intended to ensure safe and human-preference-compliant outputs, as discussed in the preceding section.

The specific reasons why these safeguards fail under certain jailbreak attacks remain insufficiently understood in the literature. A prominent hypothesis attributes these failures to two primary factors: competing objectives and mismatched generalization~\cite{wei_jailbroken_2023}. Competing objectives occur when an LLM prioritizes the “helpful” objective over the “harmless” constraint in response to a given prompt, resulting in unsafe outputs. For instance, a chatbot might be prompted with a harmful request disguised as an innocuous educational inquiry, which may be sufficient to bypass safety measures. In contrast, mismatched generalization arises when the alignment safeguards fail to cover specific unsafe queries, enabling the model to generate harmful outputs based on its pretraining data.

To investigate these hypotheses further, we introduce a novel perspective grounded in the domains of language covered during LLM training. Additionally, we propose a third factor contributing to the failure of model safeguards: robustness deficiencies. First, we define key concepts that are essential for understanding the proposed domain framework. Then, we apply this framework to extend and refine the findings of Jailbroken~\cite{wei_jailbroken_2023}, providing a more precise characterization of the conditions under which model safeguards fail.

\subsection{Domain Characterization in LLM Training}

A comprehensive understanding of model training domains begins by distinguishing between explicit and implicit variables. Explicit variables are those that are deliberately incorporated into the preference dataset to guide model alignment. At present, helpfulness and harmlessness are the primary explicit variables, although other attributes such as honesty have also been explored~\cite{ouyang_training_2022}. We assume that each response generated by an LLM can be evaluated along a continuous scale ranging from 0 to 1 for each explicit variable. For example, a maximally helpful response would score a 1 for helpfulness, whereas a response that violates ethical guidelines might score a 1 for harmfulness. Notably, harmfulness and harmlessness are distinct variables that should not be conflated, as they capture different aspects of model alignment. This framework enables the visualization of the training domains of an LLM while maintaining the independence of these variables.

Implicit variables, on the other hand, are not explicitly incorporated into the preference dataset. The presence of these variables may introduce biases, which are often shaped by the individuals curating the datasets. While employing crowdworkers from diverse cultural backgrounds may mitigate some of these biases, implicit biases may still persist. For the purposes of this discussion, we assume that these biases have been addressed during the development of the preference dataset.

\begin{figure} \includegraphics[width=\linewidth]{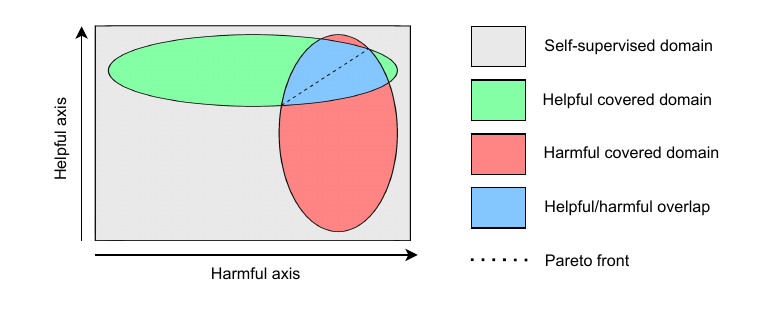} \caption{Characterization of an LLM’s training domains. The self-supervised domain encompasses the core model knowledge. The helpful and harmful domains are part of the alignment dataset. The overlap between helpful and harmful domains presents the alignment process as a multi-objective optimization task.} \label{fig:domain_analysis} \end{figure}

We now focus our attention on explicit variables. The following sections introduce the domain components. The \textit{self-supervised domain} encompasses the entire body of text utilized during an LLM’s pretraining phase. This domain can be characterized in terms of the explicit variables helpfulness and harmfulness, as illustrated in~\autoref{fig:domain_analysis}. Each sample from this domain corresponds to a model-generated text completion, and each sample is assigned respective scores for these variables.

The \textit{helpful domain}, depicted in light green in~\autoref{fig:domain_analysis}, consists of all responses within the preference dataset classified as helpful. Similarly, the \textit{harmful domain} comprises responses classified as harmful. Ideally, the model should reject queries that fall within the harmful domain. The intersection of these domains represents cases where the response exhibits both helpful and harmful characteristics.

Given the defined domains, several challenges emerge in the model alignment process. Specifically, we identify three principal challenges: mismatched generalization, competing objectives, and adversarial robustness. These challenges are described in the following section.

\subsection{Relationship to Jailbreak Vulnerabilities}

Using the domain framework described above, we can systematically characterize the weaknesses that are exploited in jailbreak attacks. The \textit{competing objectives domain} corresponds to the intersection of the helpful and harmful domains, where responses exhibit features of both objectives. Since preference learning involves multi-objective optimization, some queries in this domain may lead the LLM to prioritize helpfulness over harmlessness, thereby generating harmful responses. This vulnerability is often targeted in jailbreak attacks, as illustrated in~\autoref{sec:taxonomy}. The challenge of identifying the Pareto front—i.e., the optimal trade-off between helpful and harmful responses—is currently a focus of research~\cite{yang_metaaligner_2024, mukherjee_multi-objective_2024, guo_controllable_2024, wang_arithmetic_2024}.

A second critical vulnerability arises from \textit{mismatched generalization}. Since preference datasets cannot feasibly encompass the entire self-supervised domain, certain regions remain unaligned with human preferences. These regions, forming the mismatched generalization domain, are where model behavior becomes unpredictable. Because stochastic gradient descent does not always generalize effectively to previously unseen inputs~\cite{hendrycks_unsolved_2022}, adversaries can exploit these areas by crafting queries that bypass safety constraints.

Lastly, we introduce a third vulnerability: \textit{lack of robustness}. Certain regions within the harmful domain may not contain sufficient training examples to ensure robust alignment, leading to poor generalization. In these regions, model safeguards are more vulnerable to adversarial perturbations, enabling attackers to trigger harmful outputs.

In summary, jailbreak vulnerabilities stem from three primary domains: competing objectives, mismatched generalization, and lack of robustness. In~\autoref{sec:taxonomy}, we apply this framework to systematically categorize existing jailbreak attacks, demonstrating how various attack strategies exploit distinct weaknesses within these domains. By refining our understanding of jailbreak mechanisms, this framework lays the foundation for the development of more resilient alignment techniques for future LLMs.

\section{A taxonomy of jailbreak attacks for LLMs} 
\label{sec:taxonomy}

As previously discussed, model jailbreak refers to the manipulation of a model through prompt engineering or other techniques to elicit unsafe behavior, despite the presence of multiple safeguard mechanisms designed to prevent such actions. Effective mitigation of these threats requires a comprehensive understanding of the underlying mechanisms that enable jailbreak attacks. A systematic categorization of these attacks can provide valuable insights into their design principles and expose structural vulnerabilities within current model architectures and training methodologies.

\begin{figure}[ht]
    \begin{tabular}{ll}
        \parbox{0.49\linewidth}{
            \begin{tikzpicture}[
                    grow=right,
                    edge from parent path={(\tikzparentnode.east) -- ++(2mm,0) |- (\tikzchildnode.west)},
                    level 1/.style={sibling distance=20mm, level distance=24mm, text width=15mm},
                    level 2/.style={sibling distance=13mm, level distance=26.5mm, text width=25mm}, 
                    every node/.style={rectangle, rounded corners, draw=black, align=center},
                ]
                \node [text width=20mm, fill=blue!30] {Mismatched generalization}
                child {
                    node [fill=blue!20] {Vision \cite{gong_figstep_2023, jeong_playing_2024}}
                }
                child {
                    node [fill=blue!20] {Text}
                    child { node [fill=blue!10] {Input \cite{yuan_gpt-4_2024, yong_low-resource_2023, lemkin_using_2024, jiang_artprompt_2024, li_cross-language_2024, handa_when_2024, ghanim_jailbreaking_2024, jeong_playing_2024}} }
                    child { node [fill=blue!10] {Output \cite{huang_catastrophic_2024, zhao_weak--strong_2024}} }
                };
            \end{tikzpicture}\\\\
            \begin{tikzpicture}[
                    grow=right,
                    edge from parent path={(\tikzparentnode.east) -- ++(2mm,0) |- (\tikzchildnode.west)},
                    level 1/.style={sibling distance=20mm, level distance=24mm, text width=15mm},
                    level 2/.style={sibling distance=13mm, level distance=26.5mm, text width=25mm}, 
                    every node/.style={rectangle, rounded corners, draw=black, align=center},
                ]
                \node [text width=20mm, fill=purple!40] {Competing objectives}
                child {
                    node [fill=purple!25] {Text}
                    child { node [fill=purple!10] {Human-crafted \cite{liu_jailbreaking_2023, shen_anything_2024, yao_fuzzllm_2024, perez_ignore_2022, li_multi-step_2023, li_deepinception_2024}} }
                    child { node [fill=purple!10] {In-Context Learning \cite{wei_jailbreak_2023, anil_many-shot_2024}} }
                    child { node [fill=purple!10] {Automatic \cite{shah_scalable_2023, wu_jailbreaking_2024, chao_jailbreaking_2023, zeng_how_2024}} }
                };
            \end{tikzpicture}\\\\
            \begin{tikzpicture}[
                    grow=right,
                    edge from parent path={(\tikzparentnode.east) -- ++(2mm,0) |- (\tikzchildnode.west)},
                    level 1/.style={sibling distance=40mm, level distance=24mm, text width=15mm},
                    level 2/.style={sibling distance=15mm, level distance=26.5mm, text width=25mm}, 
                    every node/.style={rectangle, rounded corners, draw=black, align=center},
                ]
                \node [text width=20mm, fill=yellow!40] {Mixed}
                child {
                    node [fill=yellow!25] {Vision \cite{shayegani_jailbreak_2024, li_images_2024, liu_arondight_2024}}
                }
                child {
                    node [fill=yellow!25] {Text}
                    child { node [fill=yellow!10] { Noisy Comp. Obj. \cite{zhuo_red_2023, kang_exploiting_2023, ding_wolf_2024} }}
                    child { node [fill=yellow!10] { Noisy Mism. Gen. \cite{wang_robustness_2023, wang_adversarial_2023, zhang_safety_2023} }}
                    child { node [fill=yellow!10] { Jailbroken combinations \cite{wei_jailbroken_2023, liu_flipattack_2024, lv_codechameleon_2024} }}
                    child { node [fill=yellow!10] { All combinations \cite{xu_cognitive_2024, andriushchenko_jailbreaking_2024} }}
                };
            \end{tikzpicture}
        } &
        \parbox{0.49\linewidth}{
            \begin{tikzpicture}[
                    grow=right,
                    edge from parent path={(\tikzparentnode.east) -- ++(2mm,0) |- (\tikzchildnode.west)},
                    level 1/.style={sibling distance=65mm, level distance=24mm, text width=15mm},
                    level 2/.style={sibling distance=16mm, level distance=26.5mm, text width=25mm}, 
                    every node/.style={rectangle, rounded corners, draw=black, align=center},
                ]
                \node [text width=20mm, fill=teal!40] {Adversarial robustness}
                child {
                    node [fill=teal!25] {Vision}
                    child { node [fill=teal!10] { WB access \cite{niu_jailbreaking_2024, qi_visual_2024, bagdasaryan_abusing_2023, schlarmann_adversarial_2023, bailey_image_2024, gao_adversarial_2024, wang_white-box_2024} }}
                    child { node [fill=teal!10] { BB access \cite{dong_how_2023, zhao_evaluating_2023, hu_transferable_2024} }}
                }
                child {
                    node [fill=teal!25] {Text}
                    child { node [fill=teal!10] {WB noisy generation \cite{zou_universal_2023, jones_automatically_2023, lapid_open_2023} }}
                    child { node [fill=teal!10] {WB stealthy generation \cite{zhu_autodan_2024, paulus_advprompter_2024, sankar_sadasivan_fast_2024, guo_cold-attack_2024} }}
                    child { node [fill=teal!10] {BB stealthy permutation \cite{zhuo_robustness_2023, mehrotra_tree_2024, takemoto_all_2024, samvelyan_rainbow_2024, deng_masterkey_2024, yu_llm-fuzzer_2024, liu_autodan-turbo_2024} }}
                    child { node [fill=teal!10] {WB stealthy permutation \cite{liu_autodan_2024} }}
                    child { node [fill=teal!10] {BB noisy generation \cite{sitawarin_pal_2024, maus_black_2023, hayase_query-based_2024, shah_loft_2023} }}
                };
            \end{tikzpicture}
        }
    \end{tabular}
    \caption{Our proposed taxonomy for jailbreak attacks to Large Language Models. There are four groups of attacks: mismatched generalization, competing objectives, adversarial robustness and mixed attacks. \textbf{BB} and \textbf{WB} stands for Black-Box and White-Box access, respectively}
    \label{fig:jailbreak_taxonomy}
\end{figure}
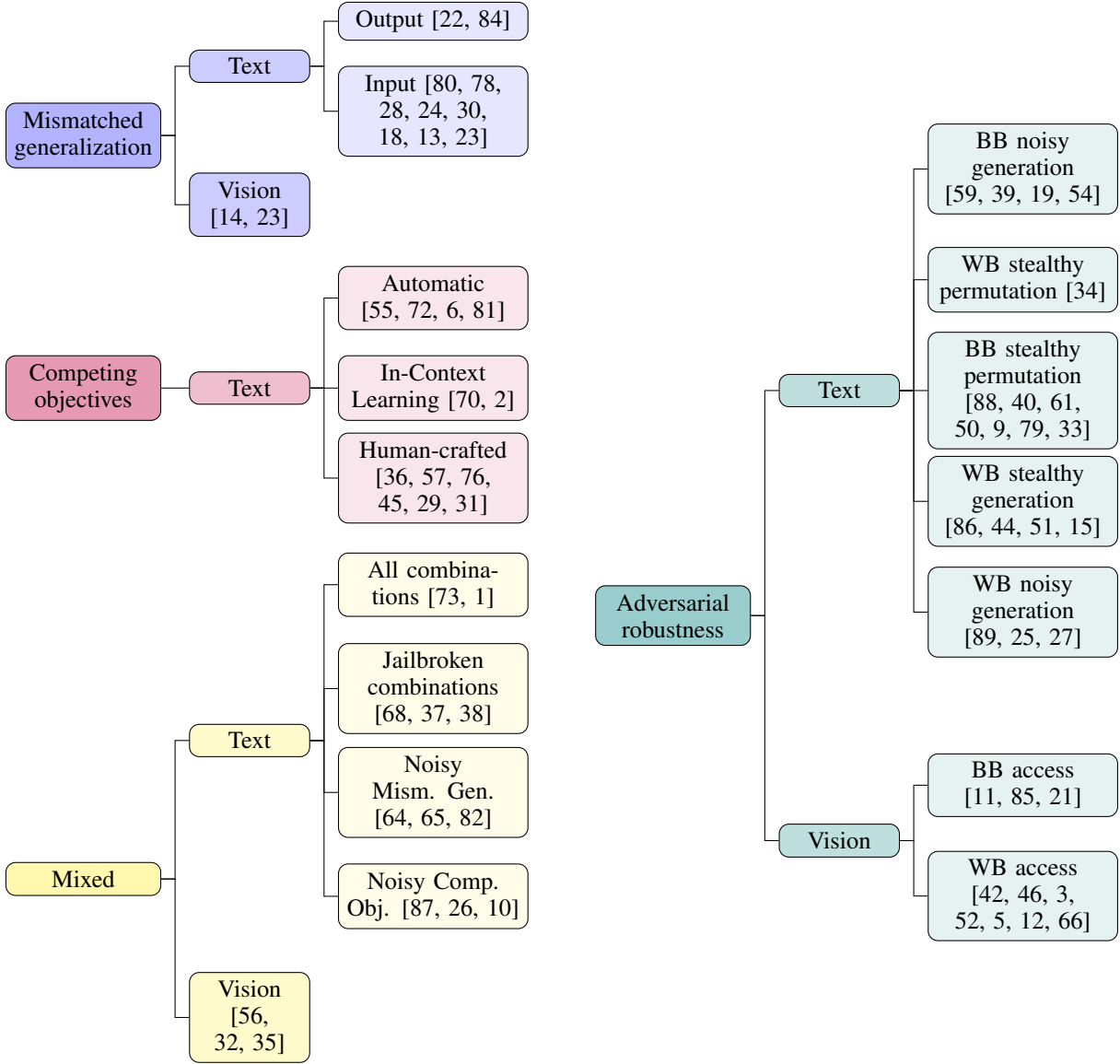

In this section, we exploit the analysis of jailbreak attacks given in \autoref{sec:domain_analysis} to propose a taxonomy of jailbreak attacks documented in the specialized literature. 
Specifically, we classify attacks according to the training domain of the LLM they exploit, namely the mismatched generalization domain, the competing objectives domain, or the lack of robustness domain. Additionally, we identify a fourth category, termed ``mixed attacks,'' which encompasses attacks that integrate techniques from at least two of the aforementioned groups. The resulting taxonomy is illustrated in \autoref{fig:jailbreak_taxonomy}.

The remainder of this section provides an in-depth analysis of these four attack categories, further subdivided based on input modality. Each subsection is classified under one of the defined categories: mismatched generalization, competing objectives, adversarial robustness, or mixed attacks.

\subsection{Mismatched generalization} 


Mismatched generalization in model alignment arises when the pre-training dataset includes specific unsafe content that is absent from the alignment dataset. Consequently, the model may generate unsafe responses when queried about such content. Exploiting this phenomenon, users can identify and target these uncovered regions to jailbreak models.

The regions or domains covered by the alignment process depend on the input modality. Recently, Large Language Models have gained the ability to process and understand not only text but also images, requiring these new vision capabilities to be considered in the alignment process. For this reason, we first analyze existing jailbreak attacks on text-only LLMs and then examine mismatched generalization jailbreak attacks on vision models.

\subsubsection{Attacks to text modality using mismatched generalization}\label{sec:mismatched_generalization:text}

\begin{figure}[tpbh]
    \centering
    \includegraphics[width=\textwidth]{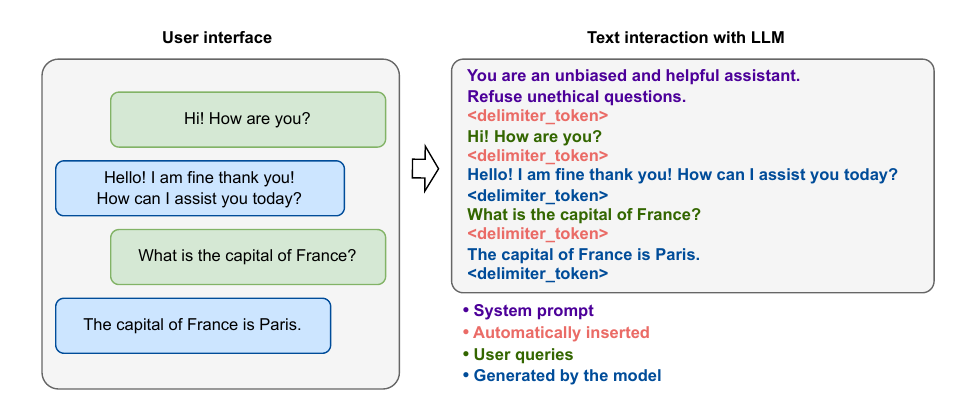}
    \caption{Prompt structure for chat models, showcasing the system prompt, the user queries and model generated tokens.}
    \label{fig:chat_prompt_format}
\end{figure}

In this section, we discuss mismatched generalization attacks on chat models, i.e., models designed to maintain a text-based conversation with the user in a friendly, helpful, and harmless manner. This conversation is typically accessible through a user interface. However, the inputs received by the chat model contain significantly more tokens than those displayed in the user interface, following a specific structure represented in \autoref{fig:chat_prompt_format}\footnote{\url{https://huggingface.co/docs/transformers/chat_templating}}. This common structure consists of three main components: the system prompt, user queries, and model-generated tokens. The system prompt, placed at the beginning of the conversation, serves as an instruction defining the model’s behavior. While not visible in the user interface, its purpose is to enhance model alignment by specifying how the model should generally respond. Following the system prompt, user queries and model-generated tokens appear sequentially, separated by a special token included in the vocabulary. For simplicity, we refer to this token as \texttt{<delimiter\_token>}. When a user submits a new query, it is appended to the conversation with a \texttt{<delimiter\_token>} following the query, signaling to the chat model that it should generate a response, which again concludes with a \texttt{<delimiter\_token>}.

These three regions of the actual prompt introduce several model vulnerabilities. The user queries region provides the most control to the user, as they can input any string, with the only restriction being the vocabulary available. Consequently, user queries are the primary focus of jailbreak attacks. If accessible, modifying the system prompt is another method of jailbreaking a model, which is why model vendors keep this prompt hidden from users. Regarding model-generated tokens, vendors impose significant restrictions, the most notable being the inability to insert tokens in the output region\footnote{\url{https://platform.openai.com/docs/guides/text-generation}}. For example, it is not possible to set an initial response and have the model continue from it. In contrast, model-generated tokens are typically mutable when using white-box access models. These models operate in a well-controlled environment where the execution code and model weights are managed by the user, allowing them to programmatically set or insert any desired token in any region.

Inspired by this conversation structure, we focus on two of these vulnerable regions, each requiring different defense strategies. These are named as input mismatched generalization and output mismatched generalization. Both are described below.

\paragraph{Input mismatched generalization} is defined as a mismatched generalization in the input prompt. That is, this occurs when the input prompt is an unsafe query not covered by the alignment dataset. A feature of this type of mismatched generalization is that defenses can be implemented both before and after the model prompt. Defenses implemented before prompting the model aim to determine whether the user prompt asks for any unsafe query. Defenses after prompting the model assess whether the model-generated content is unsafe or not. Input mismatched generalization can be defended using these two strategies, as the user prompt may be classified as harmful before the target model generates its response. Another feature is that users usually have full control over the input prompt, so the implementation of defenses is more challenging.

A common example of input mismatched generalization is the use of poorly represented languages. Specifically, translating an unsafe query into a low-represented language, directly prompting the model with the translated text, and getting back the answer in the original language could jailbreak the model~\cite{yong_low-resource_2023, li_cross-language_2024, ghanim_jailbreaking_2024}. Another possibility involves using the emergent capabilities of large language models, particularly their ability to cipher/decipher messages using simple mechanisms. More specifically, this can be achieved by prompting the model to send a ciphered message and forcing it to decode the message and follow the instructions within, leading to unsafe responses. The ciphers used include character encoding (ASCII, UTF, Unicode), as well as common ciphers such as Caesar or Atbash~\cite{yuan_gpt-4_2024, handa_when_2024}. Other examples of generalization attacks with mismatched input include the use of ASCII Art to encode unsafe keywords~\cite{jiang_artprompt_2024} or exploiting the hallucination feature of the models~\cite{lemkin_using_2024}. Finally, it is also possible to transform input queries into Out-of-Distribution queries by randomly mixing unsafe and safe words, then jailbreak the model using these transformed prompts~\cite{jeong_playing_2024}.

\paragraph{Output mismatched generalization} occurs when the mismatched generalization of the model is exploited by attacking the tokens generated by the model. Although unsafe input prompts may be included in the alignment dataset, it is possible to jailbreak the model for such prompts by modifying the model's behavior during answer generation. A key difference from input mismatched generalization is that defenses cannot be easily implemented through pre-processing techniques. Instead, vendors must account for these attacks and avoid exposing API functionalities that would allow the implementation of jailbreaking techniques under these conditions. As a simple example, if we have write-access to the text generated as output (see \autoref{fig:chat_prompt_format}), we can introduce \textit{Sure! Here is how} as model-generated tokens. When the model is asked to continue and complete the sentence, a jailbreak is achieved~\cite{zhang_safety_2023}. Even though aligned models tend to refuse harmful queries initially, once their answer prefix contains the start of a harmful response, they are likely to complete the answer in a harmful way.

Another relevant case of output mismatched generalization is the use of sampling methods to jailbreak a model. In fact, it is possible to jailbreak a model simply by modifying the current token sampling parameters. For example, if a model uses the top-\textit{k} token sampling method by default, changing the value of \textit{k} to a different value can lead to unsafe answers~\cite{huang_catastrophic_2024}.

Finally, if we have full access to the output probability distribution of the model, this probability distribution can be modified to jailbreak the model, as described in the work of~\cite{zhao_weak--strong_2024}. Let us assume we have access to an extremely capable but safe LLM. The authors of~\cite{zhao_weak--strong_2024} hypothesize that model alignment primarily modifies model behavior in the first generated tokens. That is, if aligned and unaligned models are asked to complete a partially written answer, both models are likely to produce a similar response. This can be exploited by modifying the probability distribution of the output of the capable and safe model, using the probability distribution of the weak and unsafe model. The key idea is to merge both distributions in such a way that the answer will likely start with the tokens chosen by the weak model, but progressively, the capable and aligned model will have more influence on the tokens produced, leading to an unsafe answer while retaining the capabilities of the stronger model.

Jailbreak attacks implemented through output mismatched generalization can be prevented by not exposing ways to modify or condition the model output. However, this defense cannot be implemented for white-access weights, where users can modify the model behavior in both input and output generation.



As Large Language Models were initially designed to generate text from text, these attacks focus on this modality. However, as LLMs evolve, more modalities are being implemented. For example, vision capabilities have been added to these models, introducing new attack vectors. In the following section, these techniques from the literature will be discussed in the context of mismatched generalization.

\subsubsection{Attacks to vision modality using mismatched generalization}

The implementation of new modalities into Large Language Models has enabled new jailbreak possibilities. For mismatched generalization, both the pretraining domain and the alignment domain have expanded. However, this expansion is not necessarily proportional, as adding pretraining data is typically easier than adding alignment data. This is because pretraining data is usually collected by scraping web pages or other sources, while alignment data is manually generated. For these reasons, adding new modalities to the models may increase the likelihood of jailbreaking models through mismatched generalization.

Mismatched generalization in multimodal models can be implemented in several ways. For example, it is possible to render text within an image and ask the model to read and complete the query, even if it is unsafe~\cite{gong_figstep_2023}. Another example, already presented for text-only models, is also applicable to vision modalities. Specifically, generating new images that are far from the alignment dataset (Out-of-Distribution images) can jailbreak the models~\cite{jeong_playing_2024}.


\subsection{Competing objectives} 


We say there are competing objectives for a Large Language Model when the model is prompted to accomplish multiple objectives that conflict with each other. These competing objectives typically involve a normally rejected query and some secondary objective that causes the model to accept the query. The normally rejected query is the task we are trying to accomplish, such as unsafe content generation, private data leakage, or system prompt leakage. On the other hand, the secondary objective is generally classified into different categories defined by the community~\cite{shen_anything_2024}. A well-known example of such a category is `Do~Anything~Now,' where the model is asked to ignore all ethical considerations.

The objectives indicated in the prompt depend on the modalities supported by the model. Modalities include text, vision, and other capabilities of a Large Language Model. If more than one modality is supported, the objectives can be distributed across the different modalities. For this reason, we first cover attacks on the text modality, and then explore how competing objectives could be utilized with the vision modality.

\subsubsection{Attacks to text modality using competing objectives}

The way a jailbreak prompt is built can determine the defenses that can be implemented against competing objectives attacks. Human-crafted attacks are prompts designed by the community, and defenses against these could be implemented by using rules to detect such prompts. Another way to jailbreak the models is by using their In-Context Learning capability. This capability allows the model to perform better when several examples are provided in context. One possible defense against this attack could involve extracting the examples from the prompt and evaluating their toxicity. In-Context Learning jailbreak attacks require some manual data to work. To provide a more automatic way to build the prompts, algorithms can also be designed to generate them. These are automatic methods for generating jailbreak prompts.


\paragraph{Human-crafted jailbreak attacks} are manually built prompts that include one or more secondary objectives. This allows the user to bypass the alignment of a model. These prompts are encoded as templates. A template is a query that includes one or more placeholders, allowing the user to insert an unsafe query and additional data. An example of extra data is a constraint that restricts the model's behavior. A particular challenge with this approach is the use of placeholders, as automatically inserting a query into the template might create a syntax incoherence, which could lead the model to misunderstand the query.

Human-designed prompts emerged naturally as the community began generating these kinds of prompts and posting them on the internet. These jailbreak prompts were collected from different sources and categorized into several classes, including changing the narrative style, working in virtual scenarios, and more~\cite{shen_anything_2024, liu_jailbreaking_2023}. It is common for these prompts to include a single secondary objective, but it is also possible to include more than one. Under this approach, concatenating several secondary objectives increases the probability of jailbreaking the model~\cite{yao_fuzzllm_2024, li_multi-step_2023}. Defining and merging these secondary objectives is a manual process. Instead of explicitly defining them, it is possible to make the model generate them by instructing it to iteratively produce the objectives in the chat. More specifically, it is possible to prompt the model to generate nested stories to jailbreak the model~\cite{li_deepinception_2024}. The typical task solved by these methods is the generation of unsafe content. However, other tasks, such as goal hijacking and prompt leakage, can also be targeted. This task is accomplished by using instructions that ask the model to ignore any previous instruction given, including In-Context Learning examples or any other instructions~\cite{perez_ignore_2022}.

\paragraph{In-Context Learning jailbreak attacks} leverage the emerging capabilities of Large Language Models to bypass safety checks. To do so, several examples of unsafe behavior are provided to the model to elicit the same behavior in the model's generation. We do not categorize this kind of attack as an automatic method because it requires examples of the desired behavior and a manually generated template to insert the examples and the query.

Using In-Context Learning, both an attack and a defense are proposed. The attack and defense are carried out using few-shot examples. The few-shot attack has been shown to work even if the topic of the examples does not match the topic of the query~\cite{wei_jailbreak_2023}. With the trend of increasing model context sizes, it is also possible to use many-shot examples, as these fit within the context. By generating the examples using a non-aligned Large Language Model, several examples can be created. Using these examples, it is possible to jailbreak a model~\cite{anil_many-shot_2024}.



\paragraph{Automatic jailbreak attacks} generate new prompts using automatic algorithms. Specifically, for the competing objectives challenge, these new prompts are designed to include secondary objectives to confuse the model. It is common to manually select one type of secondary objective to generate jailbreak prompts around. To the best of our knowledge, there is no research on finding these secondary objectives automatically. This would allow the creation of fully autonomous competing objectives jailbreak prompts.

There are several ways to automatically build prompts based on a specific type of secondary objective. For example, the process of prompt building can be split based on persona modulation. By creating four different stages and executing the last three using Large Language Models as content generators, new prompts can be created in a way the model behave as specific personas, such that unsafe queries are answered~\cite{shah_scalable_2023}. Another type of secondary objective is persuasion. A taxonomy of persuasion techniques focused on people is developed. Using this taxonomy, a training dataset of harmful queries is transformed by persuading using each specific category of this taxonomy. Then, a Large Language Model is fine-tuned using pairs of harmful queries and persuasion queries so that the model learns how to persuade. Using this model, it is possible to feed it with new harmful queries and jailbreak a target model~\cite{zeng_how_2024}. There are other entry points to identify secondary objectives that would jailbreak a model. One approach is to use the system prompt to find vulnerabilities. First, the system prompt is leaked using the vision capabilities of the model. A new prompt is generated by asking one model to find vulnerabilities in the system prompt. Then, this prompt is further manually refined by adding explicit secondary objectives, yielding better results~\cite{wu_jailbreaking_2024}. If the user has access to a model with the ability to modify the system prompt, jailbreak methods based on that can also be implemented. Considering this model as an attacker model, unsafe queries are used to sample jailbreak prompts focusing on secondary objectives. These generated prompts are tested on the target model, and the sampling stops whenever a jailbreak is successful or a maximum number of iterations is reached~\cite{chao_jailbreaking_2023}.

\subsubsection{Attacks to vision modality using competing objectives}

Multi-modal jailbreak attacks are an extrapolation of other jailbreak categories in this taxonomy. However, to the best of our knowledge, there is no research that combines competing objectives with vision or other modalities in Large Language Models. There are several combinations that could generate this kind of attack. For example, it might be possible to represent the main objective in the image and the secondary objectives in the text modality, or vice versa. It could also be possible to represent both competing objectives in the vision modality. We believe this is an interesting research direction that should be further explored.

\subsection{Adversarial robustness} 


Deep learning models are typically trained using supervised or self-supervised methods. Given a dataset of input-output pairs, the task of the model is to learn how to infer an output given an input. The rules to generate this association are automatically discovered by the model and encoded into the weights. These rules are not interpretable, as the number of weights and layers defining the model architecture is too large to be understood all at once. As a result, it is possible for the model training process to discover very specific rules that fit the noise and biases present in the training dataset. These specific rules may cause the model to behave differently under small perturbations to the input. While a person would not behave differently under these perturbations, the model might generate very different outputs. This challenge in deep learning is known as adversarial robustness~\cite{silva_opportunities_2020}. Large Language Models must overcome this challenge, as they are deep learning models.

Adversarial attacks depend on the modality domain. Some modalities are discrete, such as text, while others are continuous, such as vision and audio modalities. These specific characteristics heavily affect how adversarial attacks are designed. For this reason, we first analyze attacks on the text modality and then review attacks on the vision modality.

\subsubsection{Attacks to text modality using adversarial robustness}

Adversarial robustness has been widely studied in the field of computer vision. However, there is a key difference between robustness in this field and adversarial robustness in Language Models. While computer vision uses images as input in a continuous space of pixels, text inputs to Language Models are discretized. This implies that applying gradient-based methods to find adversarial samples is more challenging. Still, adversarial robustness remains the most studied method for jailbreaking Large Language Models, as its implementation builds on previous literature from text classification and computer vision. To better categorize recent jailbreak methods using adversarial robustness, we distinguish four different perspectives. Each perspective has different defenses that need to be implemented. These are model access, type of generated noise, stealthiness, and generation method. A summary of each analyzed method categorization is shown in Table \ref{tab:robustness_text}.

\begin{table}
    \centering
    \begin{tabular}{l|x{3em}x{3em}|x{3em}x{3em}|x{3em}x{3em}|x{3em}x{3em}}
        \multirow{2}{*}{Method} & \multicolumn{2}{c|}{Model access} & \multicolumn{2}{c|}{Type of noise} & \multicolumn{2}{c|}{Stealthiness} & \multicolumn{2}{c}{Search method} \\
        & WB & BB & Mutate & Suffix & No & Yes & Grad. & Alg. \\
        \midrule
        AutoDAN (gen)~\cite{liu_autodan_2024} & $\bullet$ &  & $\bullet$ &  &  & $\bullet$ &  & $\bullet$ \\
        GCG~\cite{zou_universal_2023} & $\bullet$ &  &  & $\bullet$ & $\bullet$ &  & $\bullet$ &  \\
        ARCA~\cite{jones_automatically_2023} & $\bullet$ &  &  & $\bullet$ & $\bullet$ &  & $\bullet$ &  \\
        Open Sesame*~\cite{lapid_open_2023} & $\bullet$ &  &  & $\bullet$ & $\bullet$ &  &  & $\bullet$ \\
        BEAST~\cite{sankar_sadasivan_fast_2024} & $\bullet$ &  &  & $\bullet$ &  & $\bullet$ &  & $\bullet$ \\
        AutoDAN (grad)~\cite{zhu_autodan_2024} & $\bullet$ &  &  & $\bullet$ &  & $\bullet$ & $\bullet$ &  \\
        AdvPrompter~\cite{paulus_advprompter_2024} & $\bullet$ &  &  & $\bullet$ &  & $\bullet$ &  & $\bullet$ \\
        RobustnessCodex*~\cite{zhuo_robustness_2023} &  & $\bullet$ & $\bullet$ &  &  & $\bullet$ &  & $\bullet$ \\
        TAP~\cite{mehrotra_tree_2024} &  & $\bullet$ & $\bullet$ &  &  & $\bullet$ &  & $\bullet$ \\
        SimBAja~\cite{takemoto_all_2024} &  & $\bullet$ & $\bullet$ &  &  & $\bullet$ &  & $\bullet$ \\
        Rainbow Teaming~\cite{samvelyan_rainbow_2024} &  & $\bullet$ & $\bullet$ &  &  & $\bullet$ &  & $\bullet$ \\
        MasterKey~\cite{deng_masterkey_2024} &  & $\bullet$ & $\bullet$ &  &  & $\bullet$ &  & $\bullet$ \\
        LLM-Fuzzer~\cite{yu_llm-fuzzer_2024} &  & $\bullet$ & $\bullet$ &  &  & $\bullet$ &  & $\bullet$ \\
        AutoDAN-Turbo~\cite{liu_autodan-turbo_2024} &  & $\bullet$ & $\bullet$ &  &  & $\bullet$ &  & $\bullet$ \\
        PAL~\cite{sitawarin_pal_2024} &  & $\bullet$ &  & $\bullet$ & $\bullet$ &  & $\bullet$ &  \\
        LoFT~\cite{shah_loft_2023} &  & $\bullet$ &  & $\bullet$ & $\bullet$ &  & $\bullet$ &  \\
        AdvForFoundation*~\cite{maus_black_2023} &  & $\bullet$ &  & $\bullet$ & $\bullet$ &  &  & $\bullet$ \\
        GCQ~\cite{hayase_query-based_2024} &  & $\bullet$ &  & $\bullet$ & $\bullet$ &  &  & $\bullet$
    \end{tabular}
    \vspace{3mm}
    \caption{Text-only jailbreak attacks using model robustness. Four main characteristics are represented in each column. \textbf{WB} and \textbf{BB} stands for White-box and Black-box access, respectively. \textbf{Mutate} represents changes in the text while \textbf{suffix} indicates new tokens generation. \textbf{Stealthiness} indicates if the attack generates meaningful text. \textbf{Search method} could be gradient-guided or search algorithm. (*) in method name indicates that this is not an official name.}
    \label{tab:robustness_text}
\end{table}

\begin{itemize}
    \item\textit{Model access} determines what kind of operations can be performed on the model. We distinguish two different categories: black-box/surrogate access and white-box/gray-box access. The former is typically accessed via an Application Programming Interface (API) provided by a vendor. Access to this type of model is limited, with only generated text and, optionally, some hyperparameters being accessible or modifiable. Thus, jailbreaking these kinds of models is usually more difficult. On the other hand, white-box/gray-box access involves access to much richer information, such as the computation of gradients, access to model-generated logits, and so on.
    \item\textit{Type of noise} refers to whether the input text is mutated or new tokens are generated. There are several mutation techniques, which mainly operate at the character, word, and sentence levels. Examples of character-level mutations include the addition of typos. Word-level mutations involve word substitution with synonyms, while sentence-level mutations involve text paraphrasing. It is also possible to generate brand-new tokens without altering the initial query. These new tokens are typically appended at the end of the query, and are therefore considered query suffixes. 
    \item\textit{Stealthiness} refers to whether the newly generated content is readable or not. Non-readable text can be easily detected by perplexity filters. For example, paraphrasing methods are stealthy as long as the paraphraser generates meaningful text. However, if new tokens are generated without considering stealthiness, it is likely that the generated content will be meaningless.
    \item\textit{Search method} can be either gradient-based or algorithmic-based. Gradient-based methods rely on loss optimization using gradient-descent algorithms. These methods depend on gradients. While these algorithms are typically executed to fine-tune the model weights, it is also possible to optimize the input text while keeping the rest of the model frozen. However, since gradients cannot be accessed from black-box models, search algorithms can also be used to jailbreak a model. These can be implemented using a variety of search techniques, including tree/graph exploration, random search, and others.
\end{itemize}

We categorize the literature on robustness attacks to Large Language Models based on the characteristics described above. We group the literature using the first three characteristics: model access, type of noise, and stealthiness. We do not consider the search algorithm for grouping the methods because defenses against specific search algorithms are harder to implement compared to the other categories. Specifically, we distinguish five different categories: white-box noisy generation, white-box stealthy generation, white-box stealthy permutation, black-box stealthy permutation, and black-box noisy generation.

\paragraph{White-box noisy generation} refers to white-box targeted attacks where a suffix is appended to unsafe queries, and this suffix is generally non-legible. Generating this suffix can be done in several ways. For example, if a dataset of unsafe queries and the desired beginnings of responses is available, a loss function can be optimized to generate these responses. Given an initial adversarial suffix, each one is individually optimized using gradients and a loss function. Then, random combinations of newly generated tokens are used. The best-performing adversarial suffix is selected. This process is repeated until a successful jailbreak attempt is achieved, or several iterations are reached~\cite{zou_universal_2023}. It is also possible to generate the adversarial suffix token by token. Using the same optimization process for tokens, by changing the loss function and adding an extra step to also consider the token probabilities, it is possible to make the model predict an exact match to the target string~\cite{jones_automatically_2023}. These methods are based on gradients to guide the search. However, it is also possible to use only log probabilities or cosine similarity between a target string and generated output. This algorithm is implemented as a Genetic Algorithm (GA), where the fitness function is either one of these functions instead of the gradients~\cite{lapid_open_2023}.

\paragraph{White-box stealthy generation} refers to white-box targeted attacks where meaningful suffixes are generated to jailbreak language models. While methods like GCG~\cite{zou_universal_2023} aim to generate a target string, the loss function can be complemented. Adding the likelihood of suffix tokens to the loss function increases the readability or stealthiness of the generated suffixes~\cite{zhu_autodan_2024}. Similar algorithms have been proposed. For example, algorithms based on the beam search algorithm are adapted to jailbreak models~\cite{sankar_sadasivan_fast_2024}. These last two algorithms are static algorithms. However, it is possible to use models to generate these adversarial suffixes. Given an attack model, it is fine-tuned using pairs of unsafe queries and suffixes. Then, the generalization capabilities of the attack model are used to generate new suffixes. This allows decoupling training and inference, so generating new tokens is less computationally expensive once a model is trained~\cite{paulus_advprompter_2024}. The use of energy functions is also studied. Using these, it is possible to control not only the attack's success but also the fluency or stealthiness~\cite{guo_cold-attack_2024}.

\paragraph{White-box stealthy permutation} focuses on white-box models as targets and readable query perturbations. It has been implemented using a Hierarchical Genetic Algorithm (HGA). On the first level, paragraphs are used as the population. On the second level, each paragraph is optimized at word level. The initial population is collected from manually generated prompts. This population is optimized using the HGA algorithm~\cite{liu_autodan_2024}.


\paragraph{Black-box stealthy permutation} includes attacks targeting black-box access models using readable perturbations of queries. Directly paraphrasing a prompt can jailbreak a model~\cite{zhuo_robustness_2023}. If this paraphrasing is done iteratively, the attack success rate can increase~\cite{takemoto_all_2024, yu_llm-fuzzer_2024}. More sophisticated paraphrasing steps can be taken, such as tree exploration instead of a linear search~\cite{mehrotra_tree_2024}. For more controllability over the style of the generated jailbreak prompts, several attempts have been proposed. One of them generates jailbreak prompts using a specified style among several characteristics~\cite{samvelyan_rainbow_2024}. It is also possible to autodiscover these styles or strategies using an attacker LLM~\cite{liu_autodan-turbo_2024}.

\paragraph{Black-box noisy generation} includes attacks targeting black-box access models and the generation of non-readable tokens. These methods commonly use a surrogate model to attack the target model. A surrogate model is a model that behaves similarly to the target model. It is common for this similarity to be achieved in a local area of a domain. Using these surrogate models, there are different ways to attack a black-box access target model. For example, it is possible to fine-tune a surrogate model in the local area of a target model by generating pairs of harmful queries and target model responses. Using this locally similar model, an attack is launched to generate an adversarial prompt. Then, this prompt is most likely to transfer to the target model~\cite{shah_loft_2023}. Another way to use surrogate models is by assuming that a white-box access model has a similar behavior compared to the target model. Under this assumption and the ability to compute some kind of loss on the target model, it is possible to jailbreak it~\cite{sitawarin_pal_2024, hayase_query-based_2024}. It is also possible to jailbreak a model by just using a carefully designed `black-box loss'~\cite{maus_black_2023}.

\subsubsection{Attacks for vision modality using adversarial robustness}


Multimodal Large Language Models (MLLMs) open new ways to jailbreak LLMs because new modalities might be defined in a continuous domain, as opposed to text-only jailbreak attacks. This continuous space allows attackers to use already existing methods from Computer Vision and robustness, adapting them to MLLMs. It has been particularly well-studied for Vision-Language Models (VLMs), where the additional modality is vision. This vision capability is added to the model, allowing it to interpret not just text but also images.

\begin{figure}
    \centering
    \includegraphics[width=0.45\linewidth]{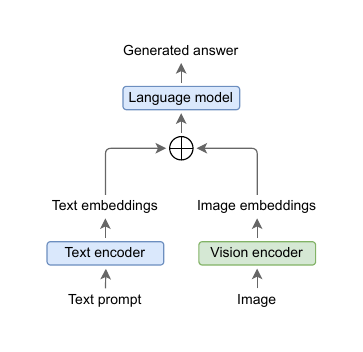}
    \caption{A common vision model architecture. Text and image embeddings are concatenated as they share a common latent space}
    \label{fig:vlm}
\end{figure}

In this section, we focus on VLMs, as they represent the most widely studied research line. Specifically, we focus on robustness attacks to these models. To understand how these attacks work, it is essential to first understand how VLMs are implemented. One common architecture for these models is illustrated in \autoref{fig:vlm}. This architecture consists of three core components: the vision encoder, the text encoder, and the main model. The vision encoder and text encoder process images and text, respectively, breaking them into tokens. Each token is then mapped into a common space known as the embedding space. The resulting embeddings are concatenated to form a single matrix, where each row represents a token (either from an image or text). These embeddings are subsequently processed by the main model, which interprets the information provided by the embeddings and generates the final output, similar to how text-only LLMs operate.

The design and implementation of jailbreak attacks to VLMs depend on the level access to the model. Similar to previous sections, we distinguish between two main access types: white-box and black-box access. White-box access provides detailed information such as gradients, logits, and the vision encoder, while black-box access offers much less information. In most cases, only the resulting text or the top-k predicted tokens are provided.

\paragraph{White-box model jailbreak attacks} often rely on existing robustness attacks used in vision models, including classifiers and object detection models. The key idea is to modify or perturb an input image to alter the model's behavior. A common process for achieving this is illustrated in \autoref{alg:adversarial_optimization}. These methods compute the gradients not from the model weights but from the input image itself. Using these gradients, it is possible to perturb the image pixels in a way that achieves the desired model behavior. This process is similar to how models are trained or fine-tuned, but instead of optimizing model weights, the image pixels are optimized. Since the pixels should not be modified beyond a certain range to keep the changes imperceptible, optimization algorithms must restrict or bound the perturbation range. This process is commonly referred to as the perturbation budget. The larger the budget or perturbation range, the easier it becomes to jailbreak the model. A common algorithm used to perturb images is Projected Gradient Descent (PGD)~\cite{niu_jailbreaking_2024, qi_visual_2024}. Other algorithms used in the literature include the Fast Gradient Sign Method (FGSM)~\cite{bagdasaryan_abusing_2023} and Auto Projected Gradient Descent (APGD)~\cite{schlarmann_adversarial_2023}. It is also possible to modify just a region of the image rather than the entire image~\cite{bailey_image_2024}. In addition to safety attacks, some studies focus on accuracy degradation in object detection tasks~\cite{gao_adversarial_2024}. While the attacks described above rely solely on the vision capabilities of the model, it is also possible to simultaneously attack both the text and vision modalities using existing methods~\cite{wang_white-box_2024}.

\begin{algorithm}
    \KwData{$I \gets Image$, $T \gets Target$, $\Theta \gets Model$, $L \gets Loss_{\Theta}(Image, Target)$, $\epsilon \gets Threshold$, $\mu \gets Step$}
    \KwResult{$I_{adv} \gets AdversarialImage$}
    $I_{adv} \gets I$; \\
    \While{$L_\Theta(I_{adv}, T) > \epsilon$}{
        $I_{adv} \gets I_{adv} - \mu \frac{d Loss_\Theta(I_{adv}, T)}{d I_{adv}}$;
    }
    \caption{Naive adversarial optimization of one image}
    \label{alg:adversarial_optimization}
\end{algorithm}

\paragraph{Black-box model jailbreak attacks} are based on the transferability of white-box access model attacks. Transferability refers to the ability to generate a perturbation for a specific surrogate model and apply this perturbed input to a target black-box model. Surrogate models in the literature include combinations of text and vision encoders, as well as complete VLMs. When using vision and text encoders as surrogate models, there are two common approaches for perturbing the image. The first approach maximizes the distance between the original image embedding and the perturbed image embedding. The second approach minimizes the embedding distance between the perturbed image and some unrelated text embedding. These methods have been explored using the CLIP model as both vision and text encoders~\cite{zhao_evaluating_2023, dong_how_2023, hu_transferable_2024}. To further enhance transferability, specific techniques from the literature, such as the Common Weakness Attack (CWA) and the Spectrum Simulation Attack (SSA), have been used~\cite{dong_how_2023}. Other surrogate models, such as complete VLMs, have also been tested~\cite{dong_how_2023, hu_transferable_2024}.

\subsection{Mixed jailbreak attacks}


Mixed jailbreak attacks combine two or more of the strategies described in previous sections: mismatched generalization, competing objectives, and adversarial robustness. To summarize briefly, mismatched generalization attacks exploit the lack of generalization in alignment that arise during the self-supervised training and alignment stages. Competing objectives attacks leverage the conflict between a harmful objective and a benign one to bypass safeguards. Adversarial robustness attacks exploit the sensitivity of deep learning models to small perturbations. The design of these attacks varies depending on the target modality. Therefore, we describe attacks on text and vision modalities in the following subsections.

\subsubsection{Attacks to text modality using a mixture of strategies}

The literature has introduced complex jailbreak attacks, but these attacks are not inherently atomic. Instead, they can be broken down into multiple atomic stages or modules, where each module employs a single strategy. We summarize the categorization strategies used in previous studies in Table \ref{tab:mixed_text}. Based on this categorization, we identify four main groups, encompassing all possible combinations of the three core strategies.

\begin{table}
    \centering
    \begin{tabular}{l|x{6em}x{6em}x{6em}}
        Method name & Mismatched generalization & Competing objectives & Adversarial robustness \\
        \midrule
        Red-Teaming*~\cite{zhuo_red_2023} &  & $\bullet$ & $\bullet$ \\
        Security-Attacks*~\cite{kang_exploiting_2023} &  & $\bullet$ & $\bullet$ \\
        ReNeLLM~\cite{ding_wolf_2024} &  & $\bullet$ & $\bullet$ \\
        ChatGPT-Robust*~\cite{wang_robustness_2023} & $\bullet$ &  & $\bullet$ \\
        advICL~\cite{wang_adversarial_2023} & $\bullet$ &  & $\bullet$ \\
        Safety-Open-Source*~\cite{zhang_safety_2023} & $\bullet$ &  & $\bullet$ \\
        Jailbroken~\cite{wei_jailbroken_2023} & $\bullet$ & $\bullet$ &  \\
        FlipAttack~\cite{liu_flipattack_2024} & $\bullet$ & $\bullet$ &  \\
        CognitiveOverload*~\cite{xu_cognitive_2024} & $\bullet$ & $\bullet$ & $\bullet$ \\
        Adaptive-Attacks*~\cite{andriushchenko_jailbreaking_2024} & $\bullet$ & $\bullet$ & $\bullet$
    \end{tabular}
    \vspace{3mm}
    \caption{Text-only jailbreak attacks using a mixture of strategies. The three main categories included in the proposed taxonomy are included. (*) in method name indicates that this is not an official name.}
    \label{tab:mixed_text}
\end{table}

\paragraph{Noisy competing objectives} includes methods that combine both adversarial robustness and competing objectives. The combination of these strategies was achieved by designing several case studies to red-team ChatGPT. Examples include role-playing or intentional word misspelling to elicit toxic behavior~\cite{zhuo_red_2023}. These case studies have also been developed using a software-oriented perspective. Several operations commonly found in programming languages have been incorporated into LLM prompts, including virtualization, variable assignment, and code obfuscation through typos. These techniques are combined to jailbreak models~\cite{kang_exploiting_2023}. A deeper integration of these strategies is also possible, where multiple operations such as paraphrasing or word reordering are applied. The resulting prompt is then embedded within a predefined scenario~\cite{ding_wolf_2024}.

\paragraph{Noisy mismatched generalization} refers to jailbreak methods that apply mismatched generalization and adversarial robustness strategies to bypass model alignment. A combination of these strategies is implemented by introducing noise at the word and sentence levels. Then, the use of Out-of-Distribution (OOD) data further enhances the jailbreak attack~\cite{wang_robustness_2023}. Another approach to incorporating OOD data is through In-Context Learning, where several perturbed examples are provided to the model to circumvent alignment restrictions~\cite{wang_adversarial_2023}. Output mismatch generalization, as described in \autoref{sec:mismatched_generalization:text}, is applied by forcing the model to begin with an affirmative response; negative words are replaced to increase the attack success rate~\cite{zhang_safety_2023}.

\paragraph{Combinations of Jailbroken strategies} utilize strategies from the Jailbroken hypothesis~\cite{wei_jailbroken_2023}. Jailbroken introduced the hypothesis of mismatched generalization and competing objectives, which are leveraged in this article. The authors designed several naive attacks using these strategies to empirically validate their hypothesis. More complex strategies have been developed. Mismatched generalization is used by changing character or word order. The probability of the model to understand the perturbed prompt is increased by applying In-Context Learning, chain-of-thoughts and role-playing techniques~\cite{liu_flipattack_2024}. Another approach involves making the model generate an unsafe query in its output and then applying an output mismatch generalization strategy. To bypass alignment restrictions and ensure the unsafe content is generated, code encryption is incorporated into the prompt, after which the model is instructed to decrypt it~\cite{lv_codechameleon_2024}.

\paragraph{Combinations of all strategies} incorporate all the strategies described in this article, including mismatched generalization, competing objectives, and adversarial robustness. For example, the existing concept of cognitive overload has been leveraged to design an attack that utilizes low-resource languages, paraphrasing, and cause-effect competing objectives~\cite{xu_cognitive_2024}. Additionally, two other attacks have been developed by combining adversarial robustness with each of the other two strategies. Adversarial robustness is implemented by randomly searching for a suffix, while mismatched generalization and competing objectives are applied using manually crafted prompts~\cite{andriushchenko_jailbreaking_2024}.



\subsubsection{Attacks to vision modality using a mixture of strategies}

Because of the addition of vision capabilities to Large Language Models, the potential for jailbreak attacks using any of the three strategies has increased. The three strategy categories (mismatched generalization, competing objectives, and adversarial robustness) have been studied for both text-only and vision modalities. The introduction of new modalities allows attackers to distribute attacks across different inputs. A common practice is encoding the harmful concept of a query within an image~\cite{shayegani_jailbreak_2024, li_images_2024, liu_arondight_2024}. This strategy is often combined with other techniques to enhance the jailbreak success rate. One approach involves perturbing benign images to reduce their similarity to harmful images, thereby bypassing toxicity filters implemented by various vendors~\cite{shayegani_jailbreak_2024}. Another technique optimizes the query using established methods alongside the harmful image, effectively embedding a mismatched generalization attack in the image while applying an adversarial robustness attack to the query~\cite{li_images_2024}. Additionally, a method for automatically generating harmful images using black-box image-to-text models has been implemented as a jailbreak technique~\cite{liu_arondight_2024}. This approach also incorporates an algorithm to increase prompt search diversity through reinforcement learning, leveraging the model’s lack of adversarial robustness.

\section{Lessons learned from the taxonomy of Jailbreak attacks for LLMs}
\label{sec:Lessons}

Based on the taxonomy of jailbreak attacks for LLMs presented in the paper, here are several lessons learned that encapsulate the key insights and implications of this work.

\begin{enumerate}
    \item Jailbreaking is not monolithic, but multidimensional. Jailbreak attacks exploit different types of vulnerabilities in LLMs. By organizing them into categories, mismatched generalization, competing objectives, lack of robustness, and mixed attacks, we gain a clearer understanding that jailbreaks arise from fundamentally distinct model weaknesses. This shifts the narrative from "how prompts are crafted" to "why jailbreaks succeed." 

    \item Alignment `gaps' are structural, not accidental. The taxonomy reveals that alignment failures are not just rare edge cases or oversights in dataset curation. Instead, they arise from inherent limitations in preference learning, especially in covering the entire training distribution and managing conflicting objectives. Therefore, existing alignment methods are structurally incomplete.

    \item Jailbreaking success is based on domain blind spots.  Attacks succeed by probing regions of the model’s behavior that are not regularized or poorly represented, such as rarely seen languages, ambiguous prompts, or adversarial perturbations. This indicates that the coverage and density of the alignment domain are crucial for safety.

    \item Mixed attacks represent the most persistent threat. Mixed attacks, which combine multiple exploit strategies, are more resilient to defenses that target only one type of vulnerability. This highlights the need for holistic defenses that account for the interaction between generalization, robustness, and conflicting optimization goals.

    \item Input and output control are equally critical. The taxonomy distinguishes between input mismatches and output manipulations, especially in multi-modal and white-box settings. This underscores that securing only the input prompt is insufficient; the model generation process and output conditioning must also be hardened.

    \item Vision and multimodal models open up new attack surfaces. As LLMs integrate vision and other modalities, new types of mismatched generalizations and adversarial vulnerabilities emerge. Safety frameworks must evolve beyond text-only scenarios to handle cross-modal exploits.

    \item Black-Box attacks are feasible and effective. Many adversarial robustness attacks in the taxonomy demonstrate that even without internal model access, attackers can succeed using transferability and surrogate models. Thus, model secrecy alone is not a sufficient defense.

    \item Prompt engineering continues to outperform defenses. The creativity and adaptability of jailbreak prompts ---especially those that take advantage of in-context learning, deception, or multistep reasoning--- suggest that defenses based solely on prompt filtering or rejection mechanisms will always fall behind.

\end{enumerate}

These observations naturally lead to several open research questions that warrant further exploration.

\begin{itemize}
    \item How can preference modeling be improved to balance safety and usability while minimizing competing objectives?
    \item What novel alignment strategies can effectively reduce mismatched generalization in a scalable manner?
    \item How can we develop more robust adversarial training methods that generalize well against new, unseen jailbreak strategies?
    \item What techniques can enhance the defensive robustness of multi-modal models against vision-based and cross-modal jailbreaks?
    \item How can LLMs autonomously detect and learn from jailbreak attempts to improve their own defenses over time?
\end{itemize}

The future research questions presented here are not isolated proposals, but rather a natural continuation of the insights distilled from our taxonomy. Each open question emerges directly from the structural vulnerabilities and patterns identified through our domain-based analysis. By forming future directions on the lessons learned, we aim to provide a cohesive roadmap to advance the alignment and resilience of the model. This integration ensures that ongoing research is both theoretically informed and practically oriented toward mitigating jailbreak risks in current and next-generation LLMs.

These research questions could be extended to provide a roadmap for advancing AI safety and improving jailbreak defenses in LLM. They could consider several aspects, categorized into different aspects of jailbreak attacks and LLM alignment, such as the following: Enhance model alignment to prevent jailbreaks, robustness against jailbreak attacks, address multimodal jailbreak vulnerabilities, adapt to emerging jailbreak techniques, ethics, and policy considerations for LLM safety. Creating a complete map of open research questions is far from the objective of the current paper. But it is an interesting open scenario and an objective for future studies to match defense analysis. 

Building on this foundation, the proposed taxonomy offers a deeper understanding of the structural vulnerabilities that make jailbreak attacks possible. By shifting the focus from surface-level prompt engineering to the underlying domain failures that models inherit during training and alignment, our framework lays the groundwork for more principled and effective defenses. Identifying these multifaceted weaknesses through targeted research and innovation will be essential for the development of safer, more robust, and trustworthy language models.

\section{Concluding Remarks}
\label{sec:conclusions}

In this work, we analyze the model alignment problem by examining the domains that emerge during LLM training through a taxonomic lens. By distinguishing between helpful and harmful domains, we introduced and formalized key concepts in jailbreak research: \textit{competing objectives} and \textit{mismatched generalization}. These insights reveal fundamental limitations of the preference learning approach to alignment. In particular, as long as competing objectives and mismatched generalization persist, jailbreak attacks will remain feasible with non-negligible probability. We also introduced the notion of a \textit{lack of robustness} region, further highlighting vulnerabilities in current alignment strategies.

Our findings suggest that existing model alignment algorithms do not fully cover the diverse corpus domain over which LLMs are trained, leaving exploitable gaps in model behavior.

To operationalize our framework, we proposed a taxonomy of jailbreak attacks categorized by the specific training domain weaknesses they exploit. This classification distinguishes attacks targeting competing objectives, mismatched generalization, adversarial robustness, and combinations thereof. By structuring the jailbreak landscape in this way, the taxonomy offers a solid foundation for evaluating, comparing, and ultimately mitigating jailbreak strategies.

Looking ahead, we emphasize the need for alignment mechanisms that inherently avoid the emergence of competing objectives. Reducing mismatched generalization will require substantially broader and more diverse preference datasets. Finally, enhancing model robustness—particularly against adversarial perturbations such as transpositions and noise—remains a key challenge for future research in developing resilient and trustworthy LLMs.

\subsubsection*{Acknowledgements} This research results from the Strategic Project IAFER-Cib (C074/23), as a result of the collaboration agreement signed between the National Institute of Cybersecurity (INCIBE) and the University of Granada. This initiative is carried out within the framework of the Recovery, Transformation and Resilience Plan funds, financed by the European Union (Next Generation).
%
%
%
\bibliographystyle{splncs04}
\bibliography{biblatex}

\end{document}